\documentclass{article}
\usepackage{graphicx} 
\usepackage[a4paper,margin=1in]{geometry}
\usepackage{newpxtext} 
\usepackage[euler-digits]{eulervm}

\usepackage{amsmath}
\usepackage[dvipsnames]{xcolor}
\usepackage[normalem]{ulem}
\usepackage{algpseudocode}
\usepackage{amsfonts}
\usepackage{graphicx}
\usepackage{adjustbox}
\usepackage{multicol}
\usepackage{multirow}
\usepackage{bbm}
\usepackage{wrapfig}
\usepackage{fancyhdr}
\usepackage{wrapfig}
\usepackage{booktabs}
\usepackage{pifont}
\usepackage{natbib}
\usepackage{soul}

\usepackage[T1,hyphens]{url}
\usepackage{hyperref}

\definecolor{mylinkcolor}{rgb}{0.6,0.3,0.3}

\hypersetup{
    colorlinks=true, 
    linkcolor=mylinkcolor, 
    citecolor=TealBlue!60!black, 
    urlcolor=TealBlue!60!black 
}

\newcommand{\cmark}{\textcolor{teal}{\checkmark}}%
\newcommand{\xmark}{\textcolor{red}{\textbf{$\times$}}}%

\newcommand{\explanation}[0]{\mathcal{E}}
\newcommand{\hexplanation}[0]{\mathcal{H}}

\newcommand{\specialcell}[2][c]{%
  \begin{tabular}[#1]{@{}c@{}}#2\end{tabular}}

\newtheorem{res-quest}{Research Question}

\definecolor{ao(english)}{rgb}{0.0, 0.5, 0.0}

\newcommand{\lmaasname}{Language-Models-as-a-Service}
\newcommand{\aasname}{as-a-Service}

\newcommand\blfootnote[1]{%
  \begingroup
  \renewcommand\thefootnote{}\footnote{#1}%
  \addtocounter{footnote}{-1}%
  \endgroup
}

\title{\lmaasname: \\Overview of a New Paradigm and its Challenges}
\author{\small{\textbf{\blfootnote{Corresponding author: emanuele.lamalfa@cs.ox.ac.uk}$^{*}$Emanuele La Malfa$^{1,2}$, Aleksandar Petrov$^{1}$, Simon Frieder$^{1}$, Christoph Weinhuber$^{1}$, Ryan Burnell$^{2}$, Raza Nazar$^{1}$}} \\ \small{\textbf{Anthony G. Cohn$^{2,3}$, Nigel Shadbolt$^{1,2}$} and \textbf{Michael Wooldridge$^{1,2}$}}}

\begin{document}

\date{
    \small{$^1$University of Oxford}, \small{$^2$The Alan Turing Institute}, \small{$^3$University of Leeds}\\%
}
\maketitle

\begin{abstract}
\noindent Some of the most powerful language models currently are proprietary systems, accessible only via (typically restrictive) web or software programming interfaces. This is the \lmaasname{} (LMaaS) paradigm. In contrast with scenarios where full model access is available, as in the case of open-source models, such closed-off language models present specific challenges for evaluating, benchmarking, and testing them. This paper has two goals: on the one hand, we delineate how the aforementioned challenges act as impediments to the accessibility, replicability, reliability, and trustworthiness of LMaaS. We systematically examine the issues that arise from a lack of information about language models for each of these four aspects. We conduct a detailed analysis of existing solutions and put forth a number of considered recommendations,
and highlight the directions for future advancements. On the other hand, it serves as a comprehensive resource for existing knowledge on current, major \lmaasname, offering a synthesized overview of the licences and capabilities their interfaces offer.
\end{abstract}

\section{Introduction}

The field of natural language processing (NLP) has undergone a profound transformation in the past few years, with the advent of (Transformer-based) Language Models (LMs)~\citep{vaswani2017attention,devlin2018bert,radford2019language}. Improved access to large models has been a fundamental facilitator of progress~\citep{wolf2019huggingface}, fueled by the scale of data and computing available to research institutions and companies~\citep{kaplan2020scaling}. 
In less than five years, the state-of-the-art models evolved from small architectures that were trainable on few GPUs~\citep{peters-etal-2018-deep,devlin2018bert} to very large and complex architectures that require dedicated data centres and supercomputers~\citep{raffel2020exploring,rae2021scaling} which are very costly to set up.
Commercial incentives have led to 
the development of large, high-performance LMs, accessible exclusively as a service for customers that return strings or tokens in response to a user's textual input -- but for which information on architecture, implementation, training procedure, or training data is not available, nor is the ability to inspect or modify its internal states
.
\par
This paradigm, known as \emph{\lmaasname} (LMaaS)~\citep{pmlr-v162-sun22e}, represents a licensing model in which LMs are centrally hosted and, typically, provided on a subscription or pay-per-use basis. Due to the performance large LMs recently attained, modern LMaaS provide a unified portal for various services that had previously been separated, from access to information that was the realm of search engines to problem-solving tools on a large number of domains such as data analysis, image generation, etc.
~\citep{romera2015embarrassingly,brown2020language,lewis2020retrieval,OpenAI2023GPT4TR} ranging from document. These services have grown rapidly and are now adopted extensively by hundreds of millions of customers.
In parallel to improving models' capabilities, the risk of malicious usage is also increasing, e.g., regarding the weaponization of biotechnologies and mass surveillance~\citep{hendrycks2023overview}. Other recent works highlight the risks of LMaaS and LMs that are not aligned with human values~\citep{bommasani2021opportunities}. This has resulted in an explosion of interest in understanding values and biases encoded by these models, intending to align the former to those of humans and mitigate the latter~\citep{ganguli2022red,liu2022aligning,santurkar2023whose}.
\par
\begin{figure}
\centering
\includegraphics[width=\linewidth]{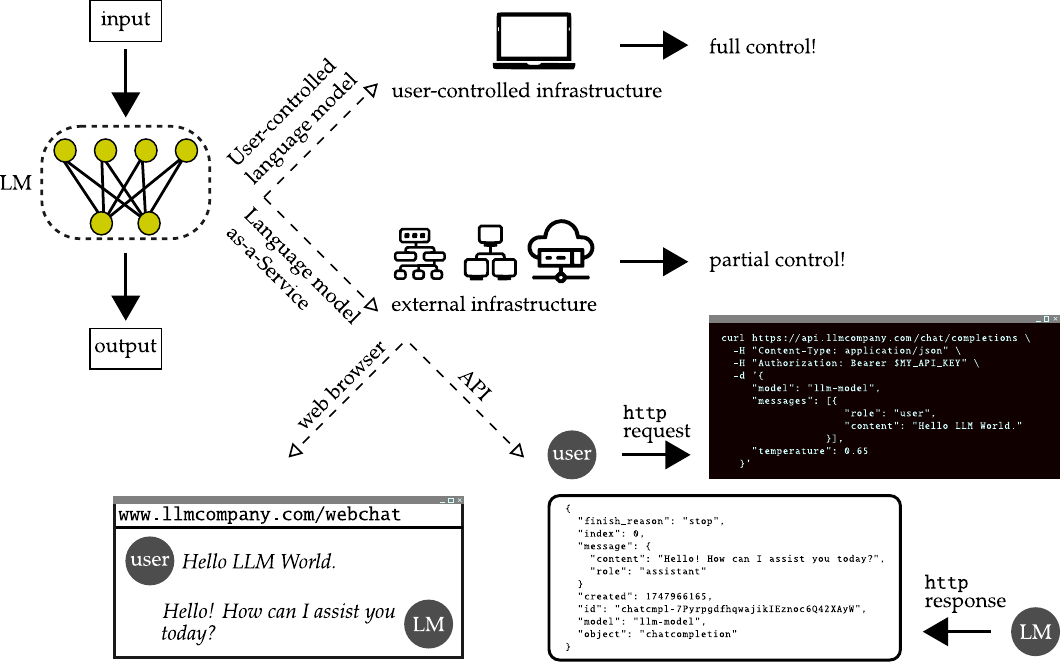}
\caption{Illustration of the difference between interacting with LMaaS and user-controlled LMs. Most LMs that are offered in full provide access to a model's internals (e.g., its weights) and list details on the training procedure and instructions on executing the models locally (the so-called model card~\citep{wolf2019huggingface}). In most cases, this allows users to run (or change) these models on the hardware of their choice. On the other hand, LMaaS are accessible through a web interface or an API (for illustrative purposes, in the diagram above, APIs resemble those of OpenAI). They are powered by LMs that run behind the scenes, typically on computational infrastructures controlled by third parties, and little information about the model is exposed to the user.
}
\label{fig:api-web}
\end{figure}

However, access restrictions inherent to LMaaS, combined with their black-box nature, are at odds with the need of the public and the research community to understand, trust, and control them better, as illustrated in Figure~\ref{fig:api-web}. 
This causes a significant problem at the core of artificial intelligence: the most powerful and economically impactful, but at the same time most risky, models are also the most difficult to analyze.
LMs are released with various licences, from open-source to more restrictive cases such as "open-weights".\footnote{\url{https://github.com/Open-Weights/Definition}} Yet, once obtained, LMs are inspectable, and the end-user can flexibly control their behaviour. On the other hand, LMaaS are accompanied by commercial licences, are mostly closed-source and partially controllable by the end-user, and are often accessible via paid subscriptions.
While some of these problems are orthogonal to the existing concerns with LMs, in general, the particularities of LMaaS exacerbate these issues or make their assessment or mitigation significantly more difficult. We group the difficulties arising from such paradigm along four categories, namely \emph{accessibility}, \emph{replicability}, \emph{reliability}, and \emph{trustworthiness}.

\begin{itemize}
    \item \textbf{Accessibility - Section~\ref{sec:accessibility}.}
        LMaaS are frequently accessible through application programming interfaces (API) or web interfaces, with free, pay-per-use, or subscription-based payment modalities. To use them, one must accept and subscribe to commercial licenses that grant companies the right to collect and use prompts to improve their model. LMs, despite some notable exceptions that we touch on in the first Section of this paper, are largely accessible and often come with permissive licenses.
        Furthermore, LMaaS access costs are often not in harmony with the socioeconomic factors of prospective users, potentially resulting in disadvantages in certain demographic layers. 
    \item \textbf{Replicability - Section~\ref{sec:replicability}.}
        LMaaS are deployed and updated in a continuous delivery/deployment regime, with legacy models frequently replaced by newer ones and depreciated altogether, possibly with little prior notice. This undermines reproducibility as one cannot evaluate withdrawn models or compare different versions.
        Replicability is further limited by the intrinsic non-determinism of LMaaS and the limited configuration options the service providers offer.
    \item \textbf{Reliability - Section~\ref{sec:reliability}.}
        Benchmarking the LMs' and LMaaS' performance 
        on various tasks and problems is how we ensure models are reliable. Benchmarking any LM incurs significant computational and human costs and is non-trivial to carry out, but for LMaaS, compared to user-controlled LMs, additional challenges occur, such as \emph{data contamination} and \emph{user contamination} (i.e., it is hard to devise samples already digested by the model and out-of-distribution test sets) and \emph{evaluating emergence} (i.e., attributing the origin of certain allegedly emergent abilities of an LMs and LMaaS).
    \item \textbf{Trustworthiness - Section~\ref{sec:trustworthiness}.}
        Models whose decision process is inspectable and interpretable in terms of elementary input-output operations are often denoted as \emph{self-explaining}. LMaaS do not belong to that class but are rather \emph{explanatory} techniques, with their explanations coming from conditioned prompts, which are handled the same way as any other query. In this sense, any explainability technique that requires access to a model's internal does not apply to them. 
\end{itemize}

These problem categories show that LMaaS require careful treatment due to their Software-\aasname{} nature (SaaS), an appropriate regulatory framework, and policies implemented by the companies that provide 
access to such services. We conclude with Section~\ref{sec:agenda}, which outlines some strategies to mitigate the most urgent issues raised by LMaaS, which we hope will help the research community and companies to make \lmaasname{} more accessible, replicable, reliable, and trustworthy. We emphasise the complementary role companies and the research community play to ensure the above values are instilled and preserved.

\section{Related Work}
Several factors have underlined the success of LMs and the consequent commercial interest and advent of the LMaaS paradigm. In this section, we provide a brief review of these developments.

\paragraph{Language Models.} 
Many computational models and approaches towards natural language understanding have been developed over the last several decades~\citep{goldberg2016primer}.
Until recently, though, their relatively poor performance limited their commercial viability.
However, the introduction of the Transformer architecture~\citep{vaswani2017attention}, inspired by previous works on attention mechanisms~\citep{bahdanau2014neural,niu2021review}, led to more efficient training of deep-learning-based LMs, which could leverage large scale text data for training.
Consequently, the performance of Transformer-based LMs quickly surpassed all previous methods~\citep{devlin2018bert}.
Large Transformer-based models, of which LMs are the most successful implementation, also exhibit zero- and few-shot behaviours: the ability to solve novel tasks with few examples or even without any \citep{chang2008importance, brown2020language}.
These behaviours are more prominent in larger models \citep{kaplan2020scaling}, which led to the development of even larger models.
This scaling was also supported by the collection of massive training datasets~\citep{shanahan2022talking,zhao2023survey}, the development of novel training and fine-tuning techniques~\citep{houlsby2019parameter,hu2021lora,bai2022training}, and improvements in computing hardware.

Parallel to that, the research community has focused on evaluating LMs and their zero- and few-shot behavior~\citep{xian2018zero,chang2023survey} as well their capability (and limitations) to solve tasks that require compositional reasoning~\citep{dziri2023faith,mccoy2023embers}. 
For example, recent works have proposed techniques to test the capabilities on unseen data points, like suites for dynamic benchmarks, behavioural testing, and out-of-domain analysis~\citep{ribeiro2020beyond,kiela2021dynabench,zhou2022domain}. 
Despite their impressive performances, state-of-the-art LMs still struggle to solve the most challenging cases of low-order tasks such as sentiment analysis~\citep{barnes-etal-2019-sentiment,barnes2021time,la2022king} or math~\cite{yang2023gpt}.\footnote{\url{https://garymarcus.substack.com/p/math-is-hard-if-you-are-an-llm-and}}
LMs have also been shown to lack robustness: they may respond incorrectly to minor variations of inputs that a model has correctly classified~\citep{sinha2021masked,wang2023robustness}.

\paragraph{\lmaasname.}
The first work we are aware of to use the term {\emph{Language-Model-as-a-Service}} is by \citet{zhao2021lmturk}, despite many publications (e.g. \citep{deng2022rlprompt,ding2022delta,dong2022survey}) citing the later work by \citet{pmlr-v162-sun22e}. LMaaS came to prominence with the advent of ChatGPT~\citep{ChatGPT} and other products developed by Google and Micosoft~\citep{thoppilan2022lamda},\footnote{\url{https://www.microsoft.com/bing}} though breakthroughs and key observations that contributed to their success date back to LMs such as GPT-3~\citep{brown2020language}.
Recently, a lot of research has focused on evaluating the performance of LMaaS on consolidated NLP datasets and benchmarks~\citep{liang2022holistic,chang2023survey,laskar2023systematic}, as well as on specialized tasks such as mathematical and spatial reasoning, symbols manipulation, and code generation~\citep{chen2018execution,kojima2022large,cohn2023dialectical,frieder2023mathematical,ray2023chatgpt,roziere2023code,shen2023chatgpt}.
These evaluations continually evolve and improve, representing a crucial component in advancing the state-of-the-art in NLP~\citep{wang2023robustness,zhao2023survey}.
LMaaS also achieve super-human performance on a variety of tasks~\citep{bubeck2023sparks}, yet fail on edge-cases that humans correctly classify~\citep{berglund2023reversal,hao2023reasoning,kocon2023chatgpt}, exhibit implicit linguistic biases~\citep{ahia2023all,bang2023multitask,petrov2023langauge} 
or are brittle to adversarial prompts~\citep{kocon2023chatgpt,shen2023chatgpt,schlarmann2023adversarial,zou2023universal}.
They also do not model uncertainty or treat ambiguity properly~\citep{liu2023we}.
We note that since the best-performing language models are the ones available solely \aasname~\citep{OpenAI2023GPT4TR}, many investigations concerning advanced capabilities of LMs cannot be decoupled from the \aasname{} platform via which the LMs are offered~\citep{frieder2023mathematical,ray2023chatgpt,shen2023chatgpt}.

\section{The LMaaS Paradigm}
We provide a high-level definition of LMs, which we then restrict to that of LMaaS. 
We then discuss how the LMaaS differs from LMs in four crucial respects, namely accessibility, replicability, comparability, and trustworthiness. 
\par
A language model defines a probability distribution over a finite string of tokens~\citep{du2022measure} and is trained to predict a symbol or token (an instance of a sequence of characters) from a finite vocabulary. 
Instead of computing the token with the highest probability over the next in a sequence, most LMs introduce diversity by employing non-deterministic sampling strategies~\citep{holtzman2019curious}.
While more capable than the deterministic counterparts, the behaviour of non-deterministic models varies according to parameters such as the \emph{temperature} or the \emph{seed}. The former is expected to make the generative process deterministic. At the same time, with the latter, we refer to a setting where all the sources of uncertainty, including the generative process, are deterministic and thus replicable.
\par
LMs and LMaaS are in-context learners~\citep{brown2020language,dong2022survey}, which refers to the ability to solve a task without changing model weights, thereby being the ideal learning approach for models provided as-a-Service through APIs or a web interface, as illustrated in Figure~\ref{fig:api-web}. 
In their simplest form, LMaaS return utterances in response to a user's prompt, both provided textually through the mentioned means of interactions. This form of interaction removes many of the possibilities of changing the models' behaviour, although commercial offerings allow some level of fine-tuning models~\citep{openai2023finetuning}.

\subsection{Accessibility}\label{sec:accessibility}
\begin{wrapfigure}{R}{0.5\textwidth}
\begin{minipage}{0.5\textwidth}
    \includegraphics[width=\linewidth]{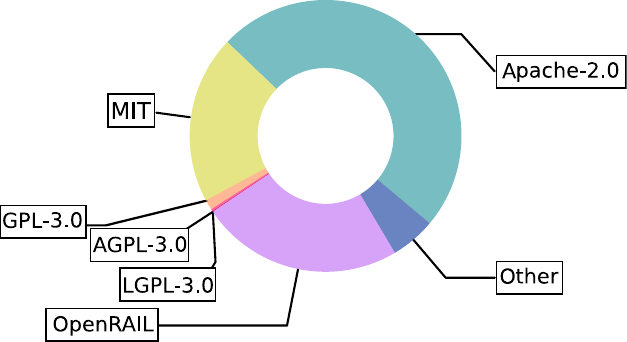}
    \caption{LMs, sourced from the \url{https://huggingface.co/models} library, grouped by how they are provided regarding software licences. Half of the LMs on Huggingface are provided with Apache-2.0 licence, followed by the MIT licence and the OpenRAIL licence, which is a new licence, specifically devised for machine-learning applications. 
    Last access 01/08/2023.
    }
    \label{fig:pie-lm}
\end{minipage}
\end{wrapfigure}

Licenses are (legal) instruments that accompany most LMs and LMaaS and govern the use and distribution of software. They define how the software can be used, modified, and distributed and outline the rights and responsibilities of the user and the software provider. Open-source and free software are widely spread philosophies that promote transparency, customizability, and community-driven development, which can enhance or limit the accessibility of software products. The landscape of LMs' licenses is variegated, as discussed in the next section and shown in Figure~\ref{fig:pie-lm}, while for LMaaS, companies mostly employ commercial licenses due to their nature as-a-Service tools. Further, we illustrate how LMaaS, delivered as pay-per-use services, intensify disparities between high-resource (e.g., English) and low-resource languages on additional components such as the tokenization process, and thus before a model is prompted.
\paragraph{LMaaS and LMs licenses.}
While some LMs require massive computational capabilities to be instantiated locally~\citep{rae2021scaling,chowdhery2022palm}, most are freely available for download, inspection, and execution on middle- to low-end machines: an instance of Alpaca-7B~\citep{alpaca} or LLaMA-7B~\citep{touvron2023llama,touvron2023llama2} can be used to do inference in full-precision, on a single GPU with 28GB of RAM.\footnote{\url{https://finbarr.ca/how-is-llama-cpp-possible/}, accessed on 26/09/2023.} 
Furthermore, approaches such as Petal~\citep{borzunov2022petals} allow users to access and fine-tune distributed LMs such as LLaMA and Falcon, leveraging a peer-to-peer network where one can contribute by sharing their computational power.
LMs licences come in different shapes and forms, as reported in Figure~\ref{fig:pie-lm}. 
Most LMs come with an MIT or an Apache license, which allow users to use, copy, modify, distribute and sell copies of the software. The MIT license does not require the source code to be available when redistributing the software unless the derived code comes in turn with an MIT license. In contrast, Apache licenses must include a copy of the license and a list of any modifications made to the original software, whether the derived product is released with the source code or in binary form. Popular LMs that are licensed with an MIT or an Apache license are respectively GPT-2~\citep{radford2019language} and BERT~\citep{devlin2018bert}. OpenRAIL is a relatively recent licensing method developed to account for cases specific to AI-powered software. It allows free access and re-distribution of its licensed material or derivatives, as long as they credit OpenRAIL and license their new creations under identical terms. On the other hand, OpenRAIL specifies use-based restrictions clauses to account for potential social costs stemming from harmful uses of openly licensed LMs. BLOOM is a popular example of an OpenRAIL licensed LM~\citep{scao2022bloom}.
Taking on a broader perspective, LMs are released through popular libraries that allow users to share machine learning models, such as Huggingface~\citep{wolf2019huggingface} or AllenNLP~\citep{gardner2018allennlp}: nonetheless, not all them fall within the definition of open source or free software~\citep{SaaS,open-free}. For example, models such as LLaMA are "available-weights", in the sense that one requests access to the model and is then granted access to the weights,\footnote{\url{https://www.alessiofanelli.com/blog/llama2-isnt-open-source}, accessed on 10/11/2023} and come with custom licenses that limit their usability. LLaMA license further forbids using its output to train other models.\footnote{\url{https://ai.meta.com/resources/models-and-libraries/llama-downloads/} section \texttt{v.}, accessed on 10/11/2023.}
\par
LMaaS, and in general Software-as-a-Service (SaaS), move the computation burden to the provider's servers though at the cost of the users no longer having direct control over the software. With capabilities that often largely surpass that of LMs~\citep{liu2023agentbench,OpenAI2023GPT4TR}, LMaaS have become a tool used daily by users and researchers. As previously mentioned, free software is software users can run, copy, distribute, study, change, and improve~\citep{SaaS} with some reasonable limitations, e.g., responsible use for OpenRAIL. LMaaS violate some or all of these principles, depending on who is offering the service, as most come with commercial licences~\citep{liesenfeld2023opening}, and cannot be run locally. The options are either paying an API provider or using a \emph{freemium} service, i.e., free for basic usage with the possibility to upgrade it with a paid subscription, offered through a web interface.\footnote{Without even counting that different models may serve different users, in the same way as LMaaS served through the API potentially differs from that hosted through the web interface} For models such as Google Bard~\citep{thoppilan2022lamda} and Microsoft Bing, prompting through API is not even supported natively (see Table~\ref{tab:lmaas-license}), yet workarounds exist: despite wide adoption, such solutions create problems regarding the liability of the software and the reliability of the results, as they might break the commercial licence under which the end user is expected to use the service.
Consequently, copying, distributing, fully studying, and changing most commercial LMaaS is impossible. 
\par 
Aligning the principles of open source to the procedures and practices of LMaaS passes through the release of their source code and the training instructions. A few institutions and companies have released LMs, such as Alpaca and BLOOM, that implement these practices~\citep{scao2022bloom,alpaca}, while some others have faced criticisms, as in the previously discussed case of LLaMA.\footnote{\url{https://opensourceconnections.com/blog/2023/07/19/is-llama-2-open-source-no-and-perhaps-we-need-a-new-definition-of-open/}, accessed on 15/10/2023.}
However, in commercial products released as-a-Service, licences become more restrictive, as companies prefer to avoid open-software methodologies~\citep{liesenfeld2023opening} to protect, via \emph{security through obscurity} their intellectual property and the derived competitive advantages. 
With the computational power distributed unevenly and concentrated in a tiny number of companies,
those with a technological, yet not computational, advantage face a dilemma. While open-sourcing their LMaaS would benefit them in terms of market exposure and contribution to their codebase by the community, releasing the code that powers a model may rapidly burn their competitive advantage in favour of players with higher computational resources~\citep{henkel2009champions,heron2013open,tkacz2020wikipedia}.

\paragraph{Access barriers to LMaaS.}
AI-powered solutions such as LMaaS will drive significant economic growth in the forthcoming years. For instance, the UK government has projected that AI could contribute to a 10\% increase in GDP.\footnote{\url{https://assets.publishing.service.gov.uk/media/5ff3bc6e8fa8f53b76ccee23/AI_Council_AI_Roadmap.pdf}, accessed on 10/11/2023.} Given the substantial economic
implications, ensuring widespread access to these models and services becomes paramount. Nevertheless, the prevalent practice of uniform pricing poses a barrier, particularly limiting accessibility for individuals and organizations in under-developed and developing areas of the world.
Consequently, the paid models offered by LLMaaS may result in considerable disparities in the economic impacts of this technology worldwide~\citep{zarifhonarvar2023economics}. Moreover, due to commercial interests, developers of LMaaS tend to focus on affluent markets. This bias is evident in the performance disparities for less commonly used languages, as noted in various studies on tokenization~\citep{ahia2023all}. Additionally, the pricing model, which charges per token or Unicode character, disproportionately affects certain languages~\citep{petrov2023langauge}. For instance, some languages incur costs up to fifteen times higher than English.
Consequently, the exclusive and paid structure of LMaaS contradicts the assertion that language models can promote global welfare and reduce social inequalities. On the contrary, they are more likely to aggravate these disparities.
A starting point to mitigate these issues is thus analyzing the impact of LMaaS and, more generally, pay-per-use artificial intelligence services as a standalone, pervasive, and disruptive technology. We call researchers from different disciplines, particularly economics and social sciences, to action to help comprehend the phenomenon from a multi-disciplinary viewpoint. For works highlighting unfair premiums paid by low-represented groups, e.g., non-English speakers, we argue that the analysis would be better informed by different groups' geographical and economic factors (for example, using informed indicators such as the monetary measure of the market value). Multi-disciplinarity is not a whim but a necessity, as solving disparities in the access to disruptive solutions such as LMaaS goes well beyond addressing their technological problems and requires implementing ad-hoc policies and governance tools.
\newline \newline
\textbf{Keywords. \textcolor{TealBlue!60!black}{API vs. web interface, free software, uneven access, premium.}} \\

\subsection{Replicability}\label{sec:replicability}
Replicability is a fundamental concept in science, serving as a cornerstone for establishing the reliability of findings. In machine learning, replicability refers to achieving the same results using the same dataset and algorithm. In this section, we discuss how LMaaS hardly meet these conditions. 
Their probabilistic nature, exacerbated by the access policies as-a-Service, the practice of deprecating models seamlessly, and their intrinsic non-determinism, affect such models even when all the sources of randomness available to the end-user are fixed.

\paragraph{LMaaS deprecation and reproducible experiments.}
In software development, continuous delivery is an approach where code changes are made to an already deployed application and released into the production environment. For SaaS solutions, this means that for a user who interacts with an LMaaS through a web browser GUI, a new version of a product can overwrite the preceding one seamlessly~\citep{ycombinator2023dumber,ycombinator2023decreased,openai2023decreased}. LMaaS, such as GPT-3.5 \& 4, Bard, and Bing, are accessible via a GUI or APIs 
, though few details are publicly available on the exact deployment strategy.
On one side, continuous delivery allows companies to deploy better models and rapidly patch bugs and newly discovered vulnerabilities; on the other, it harms the reproducibility of empirical analyses conducted on such models~\citep{chen2023chatgpt}.
For API access, one would expect clear documentation of model changes, but even in this case, reports of sudden changes in behaviour exist~\citep{eleuther2023gptmodel,sarah2020davinci} 
\par 
When a company deprecates a model, assessing and thus trusting the validity of an experiment depends only on the historical data and the consensus reached by the research community on the reliability of the benchmarking technique. 
Say that one measured, between April and May 2023, the performance of GPT-3.5 with the methodology described in~\citep{liang2022holistic}. On the 11\textsuperscript{th} of May 2023, OpenAI removed GPT-3.5-legacy from the list of available models and is adopting a similar policy for newer models.\footnote{\url{https://openai.com/blog/gpt-4-api-general-availability}, accessed on 26/09/2023.} Reproducing experiments for those models has become impossible; thus, we can only trust the correctness of the benchmarking techniques. 
On the contrary, most LMs are stored by third-party services and made available for download with permissive licences, as discussed previously in this paper. 
\par
To check if an LMaaS has been updated, one can collect some answers in response to a set of prompts, store their hash, and compare it over time to prove that a model replacement has occurred. Unfortunately, this solution is not sound: two different LMaaS can output the same answer or differ not because one replaced the other but due to their intrinsic non-determinism (see next paragraph).
Another viable solution would be to require companies to support access to legacy models for an extended period: this could happen, if not immediately after being withdrawn, after a period that would not preclude users from adopting newer versions. However, this is not without its downsides too: as newer versions patch safety and exploitation vulnerabilities (e.g., "jailbreaks" \citep{chao2023jailbreaking,liu2023jailbreaking,yao2023fuzzllm}), maintaining an "unpatched" version can thus be a method for malicious actors to continue exploiting the system. 
\par
These aspects are highly problematic, particularly when LMaaS are used as an intermediate node in a software pipeline for a downstream consumer software offering that builds on LMaaS: as LMaaS change, the quality of the downstream service will also change. One way to address these issues is to allow only access to vetted individuals and organizations, though that would limit the audit of the models.
Alternatively, providers can supply an interface for researchers to register benchmarks that re-evaluate a model at every update, like public continuous integration testing. Since API change is not documented well presently~\citep{{eleuther2023gptmodel,sarah2020davinci}}, the most viable (but costly) alternative is running permanent regression tests~\citep{chen2023chatgpt} to monitor and statistically track model performance on metrics relevant to the downstream service, akin to the daily ``GPT-4 unicorn''~\citep{dean2023unicorn}. 

\paragraph{The myth of determinism and the hardness of prompting.}
\begin{figure}
\centering
\includegraphics[width=\linewidth]{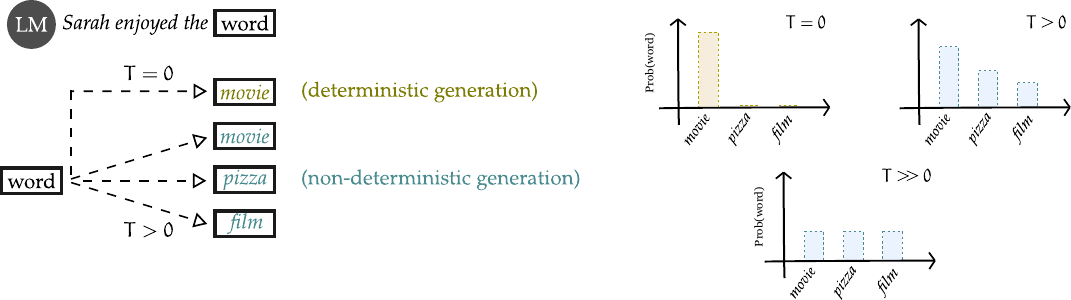}
    \caption{Setting the temperature $T$ to zero makes (typically) LMaaS deterministic, with the probability of sampling the next word that concentrates on a single word. On the other hand, for $T>0$, LMaaS become progressively more \emph{creative}, but makes the distribution over which they sample progressively more flat. 
    }
    \label{fig:detrimental-determinism}
\end{figure}
Another concern is the inherent non-determinism of most models accessible \aasname. This problem is not limited to LMaaS but extends to many generative AI techniques deployed and made available \aasname, such as diffusion-based image generators~\citep{rombach2021highresolution}.
With LMaaS, the same prompt possibly returns more than one answer, aggravating replicability. 
Models such as GPT-3.5 \& 4, when accessed via API or via the web-interface provided by cloud services such as Microsoft Azure, allow adjusting the diversity of the completion generated over a user's prompt by tuning the value of parameters such as the \emph{temperature}, as illustrated in Figure~\ref{fig:detrimental-determinism}.
In the case of GPT-3.5, the model should behave deterministically by setting the temperature to zero. Nevertheless, there have been reports that GPT-3.5 and GPT-4 are non-deterministic even for temperature zero~\citep{chann2023nondeterminism,ouyang2023llm}.\footnote{\url{https://152334h.github.io/blog/non-determinism-in-gpt-4/}, accessed on 24/08/2023.} For values greater than zero, an LMaaS samples the outputs according to a probability distribution, therefore, the sample from the distribution can be different. 
The number of utterances a deterministic LMaaS can output is a (strict) subset of those a non-deterministic LMaaS can generate: in this sense, a model gains diversity with non-determinism but could diminish the consistency of so-generated answers.
In substance, while undoubtedly helpful, controlling variables such as the temperature is insufficient to guarantee the replicability of an experiment.

Only by fixing all the randomness sources (informally, their seed) allows one to control, reproduce, and trust experiments conducted with that model. Thus, precluding access to their source code erodes trust in the end-user and possibly weakens our control of the LMaaS. 
Another related risk that is debated in literature is whether a fully deterministic model that would expose a company to the risk of seeing their models mimicked by others. While some research papers show how to train a competitive replica of an LMaaS with access solely to the output of another model or its embedded representations~\citep{mukherjee2023orca,peng2023you}, other argue that "imitation" models only apparently close the gap with powerful LMaaS~\citep{gudibande2023false}, leaving this debate unsolved for now. 
\par 
Another issue that affects transparency is that different models offer different access interfaces (e.g., via web GUI or APIs), as reported in Table~\ref{tab:lmaas-license}.
The case of GPT-3.5 \& 4 is emblematic: while the API access allows setting the temperature and other parameters, the {correspondent} web interface does not admit such options. 
With seamlessly replaced and not replicable models, the validity of any experiment, especially those that show marginal improvements over an already strong baseline, should be treated carefully. The price for validation is replicability. Any LMaaS we want to evaluate formally should provide parameters to make such analysis deterministically replicable. When that is not possible, benchmarking techniques should take non-determinism into account, e.g., via sampling strategies that explore a model output's landscape robustly to account for non-deterministic behaviours.
\newline \newline
\textbf{Keywords. \textcolor{TealBlue!60!black}{Deprecation, commercial licences, determinism, randomization.}}

\subsection{Reliability}\label{sec:reliability}
Reliability in machine learning is pivotal, signifying the consistency and dependability of a learning model's results. This concept is fundamental as it underpins the extent to which one can place confidence in the model's predictions or decisions within real-world scenarios. In the context of LMs and LMaaS, reliability encompasses models whose performance is quantifiable through metrics. These metrics should reflect the model's proficiency in accomplishing tasks that transcend mere memorization of training data and reliance on superficial patterns.
This section delves into pertinent issues for benchmarking LMs, with the LMaaS framework intensifying these assessment challenges. A primary concern is \emph{data contamination}. 
This occurs when the training data contains inputs and corresponding labels that are similar to, or exact replicas of, those utilized in the testbed. Another type of contamination, which we name \emph{user contamination}, occurs when companies collect prompts to train and fine-tune their models, and is specific to the LMaaS paradigm.
Furthermore, this section explores the "emergent behaviour" concept in LMaaS. Emergency refers to the ability of LMaaS to tackle tasks not encountered during training. Identifying such emergent behaviours necessitates a preliminary determination of whether these tasks represent novel challenges.
Lastly, we provide an overview of the ongoing dichotomy in benchmarking practices. This dichotomy lies between conducting comprehensive evaluations across multiple datasets and metrics and employing techniques that assess the meta-capabilities of such models in addressing analogous tasks.

\paragraph{Data contamination and user contamination.}
LMs are pre-trained on massive datasets that often consist of billions, if not trillions, of tokens, usually scraped as unstructured text from the web~\citep{touvron2023llama2}. Such an approach contrasts with the broader paradigm of (self)-supervised training, where models are trained and instructed on many input/output samples and tested on unseen test data. The need for massive datasets induces scarcity of test beds~\citep{van2023mitigating}, as popular ones might be available online and consumed during training due to poor scraping and data collection policies. This is the issue of \emph{data contamination}.
In this setting, performance evaluation becomes non-trivial, as it must be conducted on data not used at training time. Memorization, as the tendency of an LM to output entire sequences seen at training time, further invalidates benchmarking~\citep{ippolito2022preventing} while also being an issue from the point of privacy, by revealing personal data contained in the training dataset~\citep{carlini2022quantifying,carlini2021extracting}. 
Furthermore, models such as ChatGPT and Bard can develop answers that, while seeming correct at first glance, contain inaccurate, false, or not made-up information~\citep{alkaissi2023artificial,zhang2023language}. This phenomenon is pandemic in almost all the LMaaS and most LMs~\citep{maynez2020faithfulness}, and companies are providing their services by mentioning such issues in their licences, as reported in Table~\ref{tab:lmaas-license}.
\par
Another related issue is that of \emph{user contamination}: the correlation between the gargantuan number of parameters of an LMaaS with memorisation~\citep{biderman2023emergent,biderman2023pythia}, and their commercial licences grant companies the right to use prompts to provide, maintain, and improve their services, as illustrated in Table~\ref{tab:lmaas-license}. A model that doesn't suffer from \emph{data contamination} can be prompted with a novel test bed, that is digested by the LMaaS to invalidate any successive benchmark on it.
While some LMaaS deliver the option to opt-out of the mechanism of continual data collection, such services are not a standard practice and come with additional premium costs for the end user.
\par
There is empirical evidence that LMaaS exhibit impressive performances on low- and high-order tasks~\citep{liang2022holistic,wang2023chatgpt}, and perform expertly on many tests designed to be challenging for humans~\citep{zhang2023exploring}.
While recent developments in deep learning have significantly augmented the capabilities ofLMs and  LMaaS in manipulating symbols, the phenomenon of \emph{data contamination} might lead to overestimating their actual performance.
\par 
An illustrative case is that of the work by Zhang et al.~(\citeyear{zhang2023exploring}), which suggests that GPT-3.5 \& 4 ace entry exams at MIT. The article, subsequently withdrawn from arXiv,\footnote{\url{https://arxiv.org/abs/2306.08997}, accessed on 19/08/2023.} raises concerns related to \emph{data contamination} (among a number of other points of concern): a meta-analysis conducted on the paper suggests that results may have been contaminated~\citep{exams-mit-chatgpt}.
Issues are not limited to the case mentioned above as other researchers raised the issue of \emph{data contamination} explicitly for models such as GPT-3.5 \& 4~\citep{aiyappa2023can}: well-established benchmarks have their test set, alongside the labels, available on sharing platforms such as GitHub~\citep{jacovi2023stop}, which raises the likelihood of them being included in the model's training data.  
\par
The effort to detect and remove datasets dates back to earlier LMs such as GPT-2~\citep{brown2020language,carlini2022quantifying,mccoy2023much}. A proposed solution is to extract, from the training corpora or via testing a model's output, all the $n$-grams of a leaked test bed~\citep{brown2020language}, though more refined procedures exist~\citep{mccoy2023much}. Nevertheless, such approaches are computationally expensive and not sound: An unsuccessful result does not prove that a model has not digested (a slight variation of) it.
\par 
With classical LMs, one can detect and mitigate \emph{data contamination} by choosing unfamiliar benchmarks to assess model reliability. If one does not trust a pre-trained model, it is possible (with some effort) to retrain it locally. 
One recent proposal to mitigate \emph{data contamination} is to stop uploading test data in plain text to the internet \citep{jacovi2023stop}. This, however, is not sufficient. Models equipped with tools to browse the internet and run code (such as the GPT-4 plug-ins) could decompress the test data in plain text in their context, which can then be used for training future versions of the models. Furthermore, the proposal by \cite{jacovi2023stop} relies on an honour system: if one inadvertently exposes a test set to a model, then one should disclose it. However, knowledge of the practice and compliance would be hard to enforce.
\par
Similarly, evaluating the LMaaS can leak the test samples to future iterations of its training if the model is trained on user inputs. That is the \emph{user contamination} issue. Once a model is prompted with an input/output pair from a test bed, there is no guarantee that such a sample will not be memorized and used to train the model further (see Table~\ref{tab:lmaas-license}). A viable approach involves the deployment of models which abstain from gathering data via prompts and are rigorously trained on datasets that are open to scrutiny.

For both \emph{data contamination} and \emph{user contamination}, it is also insufficient that the \emph{exact} test set is not in the training data.
One should also ensure that \emph{the same information} is not in the training data. 
For example, even if a specific test set on solving addition problems is not ingested, one still needs to ensure that none of the individual problems has been independently developed, released online, and ingested in the training dataset.
This is especially important for simple tasks such as summing double-digit numbers, of which hundreds of exercise sheets can be found online or generated with minimal effort.

Therefore, guaranteeing that no contamination occurs is not viable.
Data handcrafted to test the model's capabilities once and then discarded, namely \emph{one-shot data}, are practically impossible to employ, as this requires human experts or specialized algorithms. Generating high-quality data with a high degree of linguistic variability is, in fact, an open problem in NLP and computational linguistics. While LMs and LMaaS have a fairly highly developed facility to generate such data~\citep{eldan2023tinystories,moller2023prompt}, they do not solve the problem~\citep{dwivedi2023so}: managing the generative process inherent in LMs and LMaaS is in fact linguistically challenging. There exists a substantial risk associated with evaluating the performance based on datasets the model may already classify with considerable accuracy, being that data generated with high confidence by the model itself.
Such a process also burdens the research community with the need to develop new datasets continuously and benchmarks, which are then immediately ingested by LMaaS providers, raising ethical concerns about the labour involved.
Therefore, ensuring that the training set does not independently have the same data is necessary.

We, therefore, believe that the \emph{data contamination} and \emph{user contamination} problems cannot be solved without the active cooperation of the model developers. However, a few interesting \emph{agnostic} approaches that estimate the probability of a sample to be part of the training data, which is assumed to be inaccessible as in the case of LMaaS, are emerging~\citep{shi2023detecting}.
For our part, we propose developing a registry for models and datasets. A researcher can check which test samples are present in the training dataset of a given model without accessing the complete training dataset directly. Then, they can omit these samples from their evaluation. For small variations of an already digested input, a solution to speed up search of similar sentences in a model training data can employ vector databases, which allow computing similarity between inputs that slightly vary~\citep{han2023comprehensive,pan2023survey}.
Such an approach also addresses the dual problem: that of a conscientious model developer wanting to avoid
the test sets of benchmarks and evaluations, which can be difficult when users might be inadvertently exposing them to the model.
A model developer can check that their training dataset does not contain test sets from the registry. Furthermore, they can dynamically check the user inputs and omit data samples that have such test samples from future training.

\begin{table}
\centering
\begin{tabular}{|l|c|c|c|c|c|c|}
\hline
\multicolumn{1}{|c|}{\textbf{Company}} &
  \textbf{Model} &
  \multicolumn{1}{c|}{\textbf{Opt-out}} &
  \multicolumn{1}{l|}{\textbf{Train}} &
  \multicolumn{1}{l|}{\textbf{Fine-tune}} &
  \multicolumn{1}{l|}{\textbf{API}} &
  \multicolumn{1}{l|}{\specialcell{\textbf{Accuracy} \\ \textbf{Disclaimer}}} \\ \hline
\textbf{AI2Lab}                                & Jurassic-1/2            & \xmark & \cmark & \cmark & \cmark & \cmark \\ \hline
\multicolumn{1}{|c|}{\textbf{Anthropic}}       & Claude 1/2                 & \cmark & $ \ $\cmark  & \cmark
& \cmark & \cmark  \\ \hline
\textbf{Cohere}                                & Command          & \cmark & \cmark & \cmark& \cmark & \cmark \\ \hline
\multirow{ 2}{*}{\textbf{Google}} & PaLM & \cmark  & \cmark & \cmark & \cmark & \cmark \\
& Bard (Powered by PaLM) & $ \ $\cmark & \cmark & \cmark & \cmark\textsuperscript{{\textdagger}} & \cmark  \\ \hline
\textbf{Inflection}          & Pi                    & \xmark & \cmark & \cmark & \xmark & \cmark  \\ \hline
\textbf{Microsoft}          & Bing                    & \xmark & \cmark & \xmark & \xmark & \cmark  \\ \hline
\multirow{ 2}{*}{\textbf{OpenAI}} & GPT-3.5/4 API & \cmark  & \cmark & \cmark & \multirow{ 2}{*}{\cmark} & \multirow{ 2}{*}{\cmark} \\
& GPT-3.5/4 web & $ \ $\cmark\textsuperscript{\ddag} & \cmark & \xmark &  & \\ \hline
\end{tabular}
\caption{
An analysis of how companies expose their LMaaS to users. Opt-out refers to rejecting data collection while still accessing the models fully. Symbols $\cmark$ and $\xmark$ indicate whether APIs and opt-out are available, and prompts are explicitly collected for fine-tuning the current LMaaS or training a new model (or product). The Accuracy Disclaimer means the relevant company does not provide any warranty as to the accuracy of the model's output. Relevant URLs to licences and access are reported in Appendix~\ref{app:data-contamination}. The Table is illustrative and thus not exhaustive: we curated a short list of models excluded from this report and the underlying motivations in Appendix~\ref{app:other-lmaas}.
\newline
\small{\textsuperscript{{\textdagger}}API access is available in Beta mode. See Appendix~\ref{app:other-lmaas} for details.} \newline 
\small{\textsuperscript{\ddag}Opt-out is the default for APIs since March 1st, 2023, while while opt-in by default for the web interface.} \newline
}
\label{tab:lmaas-license}
\end{table}

\paragraph{Evaluating emergence.} 
There have been high-profile claims that some of the most advanced LMaaS exhibit impressive emergent capabilities \citep{bubeck2023sparks,huang2023ark,singhal2023large}, i.e., capabilities not explicitly programmed or anticipated during their development that manifest when a model reaches a certain scale or complexity.
In practice, emergent abilities are behaviours that cannot be directly generalized from the training data~\citep{wei2022emergent}, and arise when models are scaled up in size and complexity~\citep{biderman2023pythia}.
For example, if a model has only seen sentiment analysis data during training, we consider its ability to solve math problems as emergent. At the same time, despite recent works showing how such properties occur in LMs~\citep{lu2023emergent}, there is reason to doubt whether these are indeed emergent abilities for LMaaS, as one cannot access the training data. Hence, one cannot evaluate the similarity of the supposed emergent ability to the abilities encoded in the training data. 

Evaluating whether LMaaS have emergent abilities is even more difficult than measuring their performance on a benchmark. While the previous section outlined ways to check whether an \emph{exact replica} of the test data is contained in the training dataset, the emergence case requires evaluating whether data for \emph{similar tasks} we want to test for are present in the training data. While a dataset and benchmark registry can be endowed with a notion of \emph{similarity measure}, finding \emph{similar} rather than \emph{exact} data is computationally challenging, when not impossible (establishing the semantic similarity between two sentences is undecidable). Therefore, for the immediate future, claims about the emergent abilities of LMaaS will likely remain highly questionable and, for LMaaS, not actionable as there is no way to access the models we want to assess (while similar tools have already been developed for standard LMs~\citep{biderman2023pythia}). 

We end with a note of caution: the existence of emergent abilities (which entail the challenges of testing them on LMaaS) is not definitively established, and in some cases, researchers discovered tasks where performances are negatively correlated with the size of the model~\citep{mckenzie2023inverse} and positively correlated with the probability of a similar prompt to be present in the training data~\citep{mccoy2023embers}. There have also been theoretical results showing that obtaining novel behaviours with prompting is challenging and that a model is good at a task indicates that it has likely seen similar or related tasks during pre-training \citep{petrov2023when}. What appear to be emergent abilities 
may be artefacts of poorly chosen evaluation metrics~\citep{schaeffer2023emergent}. This fact is exacerbated by the startling observation that an increase in performance that comes from Chain-of-Thought (CoT) prompting is still observable if the prompt has logical errors~\citep{schaeffer2023invalid}.

\paragraph{Comparing LMaaS.} 
\begin{figure}
\centering
\includegraphics[width=\linewidth]{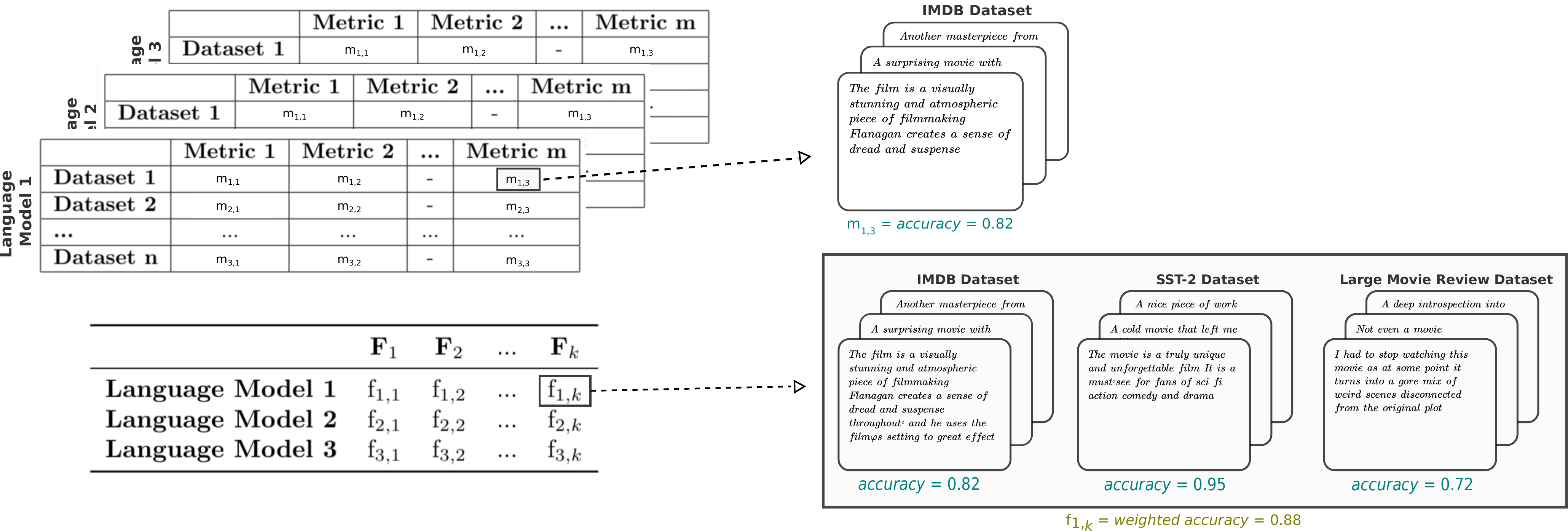}
    \caption{Extensive benchmarking techniques (top) suffer from poor inspectability as they have a cubic growth rate w.r.t. the number of models, datasets and metrics involved. On the other hand, techniques that cluster models based on dimensionality reduction and distilled latent factors (bottom) aggregate multiple datasets and/or metrics but suffer from poor interpretability.}
    \label{fig:helm-aggregate}
\end{figure}
It is not only \emph{data contamination} and \emph{user contamination} that makes it challenging to compare general-purpose LMaaS. 
While the scientific literature is moving towards multi-task, multi-models measures of performance~\citep{liang2022holistic}, or, generally speaking, towards more extensive and complex benchmarks~\citep{kiela2021dynabench,suzgun2022challenging}, the debate is still open on how to compare two models on a broader, multi-domain set of tasks~\citep{chang2023survey}. 
A model might be better at one task and lacking at another; it is thus hard to imagine a measure of performance that imposes a partial/total over LMaaS performances. Moreover, benchmarks such as BIG-Bench~\citep{srivastava2022beyond}, designed to progressively include more complex task instances, may diverge from the initial benchmarking purpose to encompass samples outside the distribution, where models exhibit failure.
\par 
Evaluating LMs and LMaaS is difficult for many reasons: with tens of models available, hundreds of metrics, and test beds, an extensive assessment of each case is expensive and necessarily time-consuming.\footnote{If the experiments reported in~\citep{liang2022holistic} were conducted solely on models with pay-per-usage costs similar to that of ChatGPT-3.5-turbo, they would cost approximately USD $18\,250$, without counting the cost of machines and engineering labour.} LMaaS add a further layer of complexity championed by \emph{data\text{/}user contamination}, model replacement, and non-determinism. If \emph{contamination} invalidates results by boosting performances on memorized test beds, experiments conducted at different time-frames do not reflect the current capabilities of a model, as a model may have been substituted in the meantime with a different version whose performances varied considerably~\citep{chen2023chatgpt}. Non-determinism, unless tamed as suggested in Section~\ref{sec:replicability}, makes point-wise comparisons uncertain and subject to larger numerical deviations.
\par 
We observe a tension in the recent literature between benchmarking via point-wise metrics on a large number of models and scenarios~\citep{liang2022holistic,srivastava2022beyond,zhong2023agieval}, and evaluations where models are aggregated \emph{post-hoc} based on latent factors that capture variations over comparable tasks~\citep{burnell2023revealing}. Such methods are prone to illusory correlation, i.e., tasks are grouped and thus considered similar based on the LMaaS performances.

An illustrative example of the difference between extensive benchmarking and latent factor analysis is illustrated in Figure~\ref{fig:helm-aggregate}: approaches such as HELM~\citep{liang2022holistic} (top), extensively benchmark LMaaS for different metrics and settings, while dimensionality reduction methods (bottom) aggregate LMs and LMaaS based on tasks where they performed similarly, yet the so-computed latent factors can be hard to interpret. 
\par 
In conclusion, rigorous benchmarking requires mitigating \emph{data\text{/}user contamination}, non-determinism, and most of the previous issues mentioned in the paper, thus representing a long-term challenge on which the research community and the model providers must necessarily collaborate. 
\newline \newline
\textbf{Keywords. \textcolor{TealBlue!60!black}{Data contamination, emergent behaviours, extensive benchmarking, latent factors analysis.}}

\subsection{Trustworthiness}\label{sec:trustworthiness}
\begin{figure}
\centering
\includegraphics[width=\linewidth]{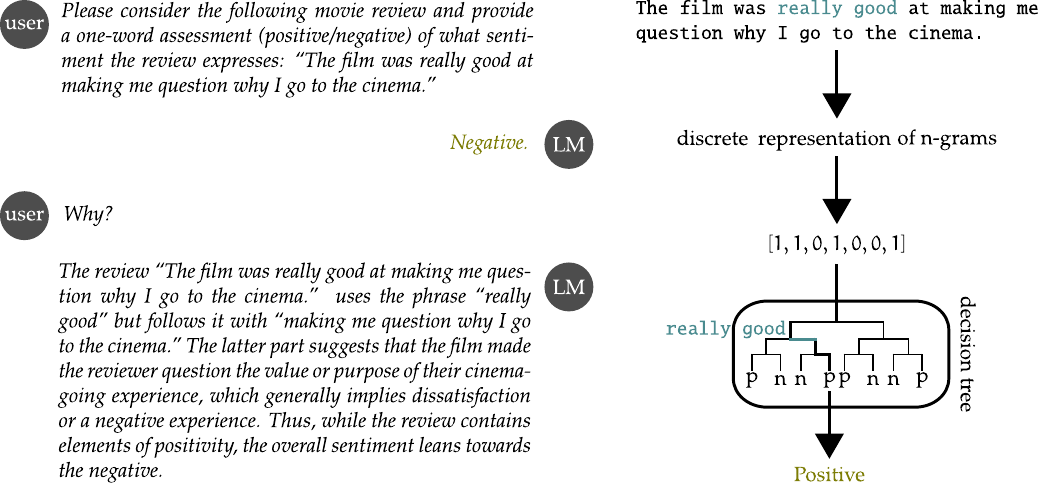}
\caption{LMaaS can be used to generate \emph{explanations}. On the left is a real interaction with GPT-4 (dated September 21, 2023), asked to solve a sentiment analysis task. In this instance the reasoning is sound, but we have no guarantee of correctness. On the right, the workings of a fictitious decision tree to solve the same sentiment analysis task are illustrated on text represented as $2$-grams: While it can misclassify the input as ``positive'', after having in represented the input as vectors, it is possible to trace the reason for the misclassification back to the existence of the $2$-gram ``really good'', which affected some of the decisions that were taken. In this sense, techniques like decision trees are \emph{self-explaining}, as they embed explanations from which one can derive the model's decision-making process.}
\label{fig:explanatory-self-explaining}
\end{figure}
LMs and LMaaS have progressively grown in sophistication up to the degree that one can query them to produce an explanation for their output~\citep{alvarez2018towards,zini2022explainability}: we refer to such models as \emph{explanatory neural networks}~\citep{alregib2022explanatory}, to distinguish them with \emph{self-explaining}, which concerns another family of explainability techniques~\citep{rudin2019stop}. 
\emph{Self-explaining} methods produce explanations that make the model's decision process transparent, as in the case of decision trees, where one can extract an intelligible rule from the branches a model follows to make its decision. 
LMaaS prompted for an explanation cannot provide insights into a model's decision process: an explanation for an LMaaS is, in fact, just a conditional prompt over a model's previous interaction, as sketched in Figure~\ref{fig:explanatory-self-explaining}. This problem also affects standard LMs, though having access to their internal states offers deeper introspection of their decision processes~\citep{azaria2023internal}.
LMaaS internals accessible through API or web interface prevents introspection and limits the explanation of a model's decision. 
An explanation only enforces trust in a model if we trust the 
explanation itself: for example, if correctness guarantees accompany it. An explanation for an LMaaS is just a conditional prompt over a model's previous interaction; hence, there is no basis to trust it.
\par 
Popular prompting techniques methods such as \emph{Chain of Thoughts}~\citep{wei2022chain} (CoT) and its variations~\citep{zhang2022automatic,besta2023graph,paranjape2023art,yao2023tree}, are employed to enhance an LMs' performance by asking a model to \emph{reason step by step and write it thoughts} when producing the answer to a problem. This step-by-step reasoning improves LMaaS performances and adds to their explainability~\citep{kojima2022large}. However, the improved performance might be due to the increasing complexity of the model~\citep{feng2023towards} and does not prevent an LMaaS from generating untrustworthy rationales~\citep{turpin2023language}. Nor do prompts have to be logically correct to increase the performance (see~\citep{schaeffer2023invalid}), as we noted in Section~\ref{sec:reliability}.
\par
Explicitness, faithfulness, and stability are three common desiderata of explanations~\citep{alvarez2018towards,maynez-etal-2020-faithfulness,li2023evaluating}. General-purpose LMaaS produce answers that maximize explicitness: in this sense, an explanation, which for an LMaaS is a conditioned prompt, will be immediate and understandable.
On the other hand, faithfulness, as the relevance of each input variable (in the case of LMaaS, each token) for the model's decision, and stability, as the consistency of an explanation for slight input variations, are not -- though some works are moving in that direction~\citep{huang2023can,lanham2023measuring}.
\par 
We argue that faithful explanations should include the sufficient causes that led a model to output an answer to a specific prompt~\citep{darwiche2020reasons}, while stability is closely connected to robustness. 
The research literature has provided overwhelming evidence that explanations for machine learning models are not sufficient to imply a model's prediction and are highly sensitive to slight input variations~\citep{ignatiev2019abd,marques2022delivering,izza2023delivering} unless explicitly trained with that objective.
Including, but not limited to, safety-critical applications, we thus advocate for methods that explicitly embody robustness guarantees, with the double intent to provide security to the end-user and not to erode their trust. Though certificates of optimality and robustness do not scale beyond small-scale models~\citep{la2021guaranteed}, we champion approximate methods and probabilistic guarantees.
Formally, for an input prompt $x$ and an LMaaS output $y$, an explanation $\explanation$ for $(x,y)$, and a set of axioms $\hexplanation$, the decision of the LMaaS entails $y$ if and only if $\explanation, x \models_{\hexplanation} y$. Such a property, combined with invariance to similar inputs and the explicitness mentioned above (which can be enforced at training time by algorithms such as reinforcement learning through human feedback and its variations~\citep{bai2022training,gulcehre2023reinforced}), can align the explainability of an LMaaS to the desiderata the research community is recommending for models that inherently suffer of poor interpretability.
In conclusion, while some works are moving in the direction of grounding explanations with external knowledge~\citep{mei2023foveate,ohmer2023evaluating}, we must develop strategies beyond recursive prompting that provide formally guaranteed and unbiased introspection of a model's decision landscape~\citep{bills2023language}, and integrate the training process with faithful \emph{post-hoc} methods~\citep{krishna2023post}.
\newline \newline
\textbf{Keywords. \textcolor{TealBlue!60!black}{Explicitness, faithfulness, stability, explanatory, self-explaining.}}

\section{Mitigating LMaaS Issues: a Tentative Agenda}\label{sec:agenda}
This work demarcates four aspects that differentiate LMaaS from LMs: accessibility, replicability, reliability, and trustworthiness.
These issues affect our ability to understand the capabilities and limitations of LMaaS, which hundreds of millions of users use daily. We, therefore, need to work as a community to find solutions that enable researchers, policymakers, and members of the public to trust LMaaS. Below, we propose a path forward and highlight challenges that must be addressed to mitigate these issues.

\paragraph{Accessibility.} To enhance accessibility, companies should release the source code (or at least detailed model cards) of the LMs that power their LMaaS. While licences that prevent free commercial usage wouldn't be enough to guarantee that companies can retain their competitive advantage (other companies could leverage their findings and larger computational infrastructures), we recommend that the source code (or very comprehensive model card) of LMaaS should at least be available to auditors/evaluators/red teams with restrictions on sharing.
Accessibility would be further enhanced by companies that release their LMs in different \emph{sizes}, as in the case of Alpaca LLaMA and Pythia, so that researchers with access to limited computational facilities can still experiment with their models. 
Regarding imbalance across languages, fair tokenizers and pay-per-token access policies can spread the usage of LMaaS among economically disadvantaged and low-resource language customers. 
We also need to assess and quantify the gap disadvantaged users and countries with limited technology access face when accessing LMaaS. Doing so can provide insights and techniques for companies and policymakers to mitigate unfair treatment.

\paragraph{Replicability.} Replicability requires that LMaaS are not taken offline when a newer version is deployed. Providing access to such newer models is certainly not remunerative for companies. Furthermore, releasing legacy {LMaaS} under permissive, if not open-source, licences is a strategy that, from a commercial perspective, likely damages providers despite allowing them to benefit from the research community discovering and reporting biases, bugs, and vulnerabilities. 
The research community would benefit from old models made available to researchers for as long as possible, with companies that warn clearly \textit{at a minimum} before updates and give lead time before deprecating old versions to enable the completion of experiments and replication efforts. At a minimum, all the parameters that make up a model should be hashed, and a log of ``model commits' should be offered to the user by model maintainers as the maintainer updates the model. The benefit is that specific user interactions with the models (in particular, benchmarks researchers make) can be matched to model commit hashes.
Companies should also offer the option to make a model's behaviour fully deterministic. We do not argue that all models released should be deterministic but that we prefer them for scientific purposes over their non-deterministic counterparts. In this sense, the scientific community, in terms of journals and conference venues, should discourage the usage of models that do not meet adequate replicability requirements. We should also develop methods for evaluating non-deterministic LMaaS under uncertainty over their seed and temperature parameters.

\paragraph{Reliability.} \emph{Data contamination} can be addressed by two complementary approaches: on the one hand, LMaaS should state clearly the datasets on which they have been trained, similarly to model cards for LMs. The industrial and research communities can collaborate to jointly develop fast indexing techniques to assess whether a model has digested an input or a slight variation. Models that collect prompts from interaction with the users should not be dismissed, but the research community, in the absence of tools to inspect whether a prompt from a test bed has been collected, should discourage their usage for reporting purposes (e.g., benchmarking).
Concerning benchmarking, we argue that the research community should study and develop benchmarking methodologies that search for and test latent factors that can explain performances across tasks while avoiding \emph{post-hoc} methods to maintain a sufficient degree of interpretability. 
As most efforts are currently directed towards the development of extensive testbeds such as HELM or BIG-Bench, which suffer from poor respectability and possibly test on out-of-distribution data, we think that more can be done in the direction of methodologies that group LMaaS and LMs by metrics and datasets, enabling interpretable model comparisons. 

\paragraph{Trustworthiness.} Finally, faithfulness and stability should be embodied in LMaaS and explainability tools that make their behaviour intelligible. 
Faithfulness is achievable through grounding and sufficiency. Despite different connotations, both terms refer to the elements of a prompt that imply a model's decision. In contrast, robustness and invariance to slight prompt variations can enable stability. Formal methods and robustness 
have already developed a rich corpus of literature from which the community should draw inspiration. The long-term objective is to deploy applications powered by LMaaS and LMs that can serve in safety-critical settings. This can instil trust in the reasons behind a model's decisions, and this is an ongoing research effort.

\section*{Acknowledgements}
ELM is supported by the Alan Turing Institute. AP is supported by the EPSRC Centre for Doctoral Training in Autonomous Intelligent Machines and Systems (EP/S024050/1). AGC is supported by the Economic and Social Research Council (ESRC) under grant ES/W003473/1; and the Turing Defence and Security programme through a partnership with the UK government in accordance with the framework agreement between GCHQ and the Alan Turing Institute.

\clearpage 
\bibliography{biblio}
\bibliographystyle{acl_natbib}

\clearpage
\appendix

\appendix

\section{Commercial Licences for Common LMaaS and Opt-Out Forms}\label{app:data-contamination}
AI2Lab \\
\url{https://www.ai21.com/terms-of-use} \\
\url{https://www.ai21.com/privacy-policy} \\ \url{https://studio.ai21.com/privacy-policy}
\newline \newline
Anthropic \\
\url{https://console.anthropic.com/legal/terms} \\
\url{https://console.anthropic.com/legal/privacy} \\ 
\url{https://support.anthropic.com/en/articles/7996868-i-want-to-opt-out-of-my-prompts-and-results-being-used-for-training-models}
\newline \newline
Cohere \\ 
\url{https://cohere.com/saas-agreement}
\newline \newline
Google \\
\url{https://support.google.com/bard/answer/13594961} \\
\url{https://support.google.com/bard/answer/13594961?hl=en} \\
\url{https://www.googlecloudcommunity.com/gc/AI-ML/Google-Bard-API/m-p/538517} - Discussion on how to access Bard via API. \\
\url{https://cloud.google.com/vertex-ai/docs/generative-ai/learn/responsible-ai} \\
\url{https://docs.google.com/forms/d/e/1FAIpQLSdUwCF62JRg8rVYh5IaN7VWwIrLtWbxtcQDRC97zbzoq54bfg/viewform} \\
\url{https://blog.google/technology/ai/an-update-on-web-publisher-controls/} \\
\url{https://myactivity.google.com/product/bard/controls?utm_source=help&pli=1} - need to be logged in.
\newline \newline
Inflection \\
\url{https://pi.ai/profile/policy} - Licence available after log-in. \\
\url{https://pi.ai/profile/terms} - Licence available after log-in.
\newline \newline
Microsoft \\
\url{https://privacy.microsoft.com/privacystatement}
\newline \newline
OpenAI \\
\url{https://openai.com/policies/terms-of-use}, \url{https://help.openai.com/en/articles/7730893-data-controls-faq} \\
\url{https://platform.openai.com/docs/models/how-we-use-your-data}
\newline \newline 
All the URLs listed above have been accessed last time on 30/07/2023. 

\section{Other LMaaS}\label{app:other-lmaas}
A list of LMaaS we decided not to include in our evaluation and the reasons behind it:
\newline 
\noindent\textbf{GitHub Copilot X}: the model is built to support programmers writing code.\footnote{\url{https://github.com/features/preview/copilot-x}}
\newline 
\noindent\textbf{HuggingChat}: the LMaaS is powered by LMs such as LLaMA, Falcon, etc.\footnote{\url{https://huggingface.co/chat/}}
\newline 
\noindent\textbf{Perplexity AI}: the model is powered by ChatGPT and Bing.\footnote{\url{https://golden.com/wiki/Perplexity_AI-X9D5GWB}}
\newline 
\noindent\textbf{Quora Poe}: it provides access to a wide range of LMs and LMaaS that can be prompted, yet none of these models is owned or has been developed by Quora.\footnote{\url{https://poe.com/}}
\newline \newline 
All the URLs listed above, except for Inflection, were accessed last time on 15/10/2023. Links regarding Inflection were accessed last time on 20/10/2023.

\end{document}